\documentclass[runningheads]{llncs}
\usepackage[T1]{fontenc}
\usepackage{graphicx}
\usepackage{amsfonts,amssymb}
\usepackage{caption}
\usepackage{color}
\usepackage{epstopdf}
\usepackage{tabularx,threeparttable}
\usepackage{array}
\usepackage[hidelinks]{hyperref}
\usepackage{graphicx}
\usepackage{subfigure}
\usepackage{amssymb, amsmath, bm}
\usepackage{mathtools}
\usepackage{mathrsfs}
\usepackage{booktabs}
\usepackage{multirow}
\usepackage{cite}
\usepackage{url}
\usepackage{color}
\usepackage{dsfont}
\usepackage{enumitem}
\usepackage[table,xcdraw]{xcolor}
\usepackage{hyperref}
\usepackage{marvosym}
\hypersetup{
    colorlinks = true, 
    citecolor = blue 
}

\begin{document}

\title{Anatomical Structure-Guided \\ Medical Vision-Language Pre-training}

\author{Qingqiu Li\inst{1} \and
Xiaohan Yan\inst{2}\and
Jilan Xu \inst{1}\and
Runtian Yuan \inst{1}\and
Yuejie Zhang \inst{1}\textsuperscript{(\Letter)} \and \\
Rui Feng \inst{1}\textsuperscript{(\Letter)} \and 
Quanli Shen \inst{3} \and
Xiaobo Zhang \inst{3} \and
Shujun Wang \inst{4,5}
}
\authorrunning{ }

\institute{Fudan University \quad 
\and Tongji University \\
\and Children’s Hospital of Fudan University \\
\and The Hong Kong Polytechnic University\\
\and Research Institute for Smart Ageing}

\maketitle     % typeset the header of the contribution

\begin{abstract}
Learning medical visual representations through vision-lang\\uage pre-training has reached remarkable progress. Despite the promising performance, it still faces challenges, i.e., local alignment lacks interpretability and clinical relevance, and the insufficient internal and external representation learning of image-report pairs. To address these issues, we propose an \textbf{A}natomical \textbf{S}tructure-\textbf{G}uided (ASG) framework. Specifically, we parse raw reports into triplets <\emph{anatomical region}, \emph{finding}, \emph{existence}>, and fully utilize each element as supervision to enhance representation learning. 
For anatomical region, we design an automatic anatomical region-sentence alignment paradigm in collaboration with radiologists, considering them as the minimum semantic units to explore fine-grained local alignment. For finding and existence, we regard them as image tags, applying an image-tag recognition decoder to associate image features with their respective tags within each sample and constructing soft labels for contrastive learning to improve the semantic association of different image-report pairs. We evaluate the proposed ASG framework on two downstream tasks, including five public benchmarks. Experimental results demonstrate that our method outperforms the state-of-the-art methods.

\end{abstract}

\section{Introduction} \label{introduction}
In recent years, vision-language pre-training (VLP) has achieved remarkable success\cite{clip, align, blip, align1}. 
These models are trained using millions of web image-text pairs by matching images to their corresponding captions without the need for manual labels. In the medical domain, this paradigm also gains increasing attention. The models trained on paired image-reports benefit a broad spectrum of downstream medical image understanding tasks.
Among them, ConVIRT\cite{convirt} first used contrastive learning as a proxy task for biomedical data processing. Building upon this, GLoRIA\cite{gloria} and MGCA\cite{mgca} further employed local-level alignment to bolster model performance. Moreover, MedKLIP\cite{medklip} and KAD\cite{kad} incorporated additional domain knowledge to guide better representation learning. MRM\cite{mrm} encouraged the model to pay attention to low-level features through mask reconstruction, enhancing model's utility in downstream dense prediction tasks. 

\begin{figure}[t]
    \centering
    \includegraphics[width=\textwidth]{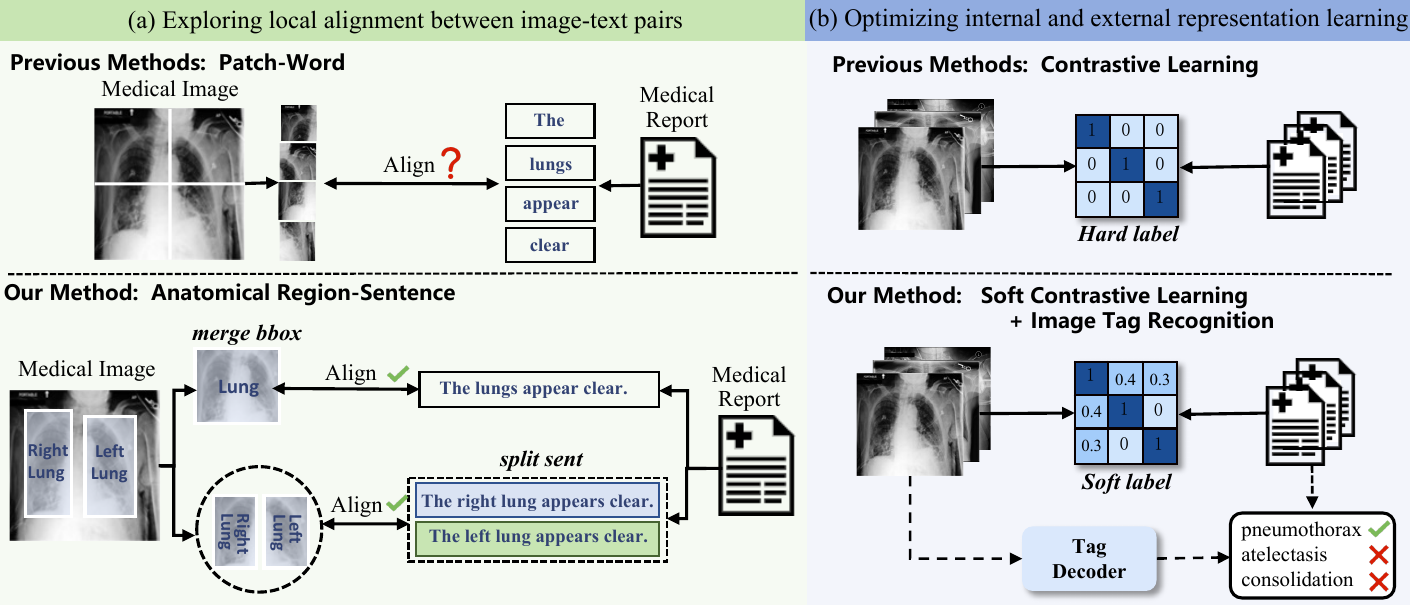}
    \caption{
    Two limitations of existing methods: (a) lack of interpretability and clinical relevance and (b) insufficient representation learning of image-report pairs; and our corresponding improvement.
    }
    \label{fig:limitation}
\end{figure}

Despite the advancements made in the medical VLP scenario, we identify two inherent limitations of existing methods that remain unresolved. \textbf{1) Lack of interpretability and clinical relevance.} Previous methods \cite{gloria, mgca} explored local alignment between image-text pairs at patch-word level or patch-sentence level, which lacks semantic and clinical correspondence. 
For example, local patches extracted from X-rays represent neither lesion area nor anatomical organs.
In addition, decomposing the whole sentence into individual words results in a substantial loss of semantic context. 
Consequently, seeking alignment between clinically irrelevant visual patches and semantically unrelated words diminishes interpretability and disturbs the model optimization. \textbf{2) Insufficient representation learning of image-report pairs.} Compared to natural image captions, medical reports are generally longer and contain richer medical knowledge, posing additional challenges for comprehensive report understanding. However, most previous approaches\cite{gloria, mgca} still rely on simple encoding of raw reports to extract text features, which fails to fully understand the entire report and capture the key sections describing lesions. KAD\cite{kad} and MedKLIP\cite{medklip} addressed this issue by using entity recognition tool to obtain structured reports, thereby improving the supervision of text for significant image representations. Nevertheless, they neglected the semantic analysis between different pairs and employed hard labels in contrastive learning, leading to a considerable number of false negatives.

In this paper, we propose a novel \textbf{A}natomical \textbf{S}tructure-\textbf{G}uided (ASG) framework to introduce anatomical knowledge into medical VLP, thus achieving clinically reliable representation learning. We parse raw reports into triplets <\emph{anatomical region}, \emph{finding}, \emph{existence}>, and utilize each element as supervisory information.
Specifically, under the guidance of radiologists, we design an automatic anatomical region-sentence alignment paradigm, which aligns with radiologists' reading workflow and enhances interpretability. Furthermore, we simultaneously focus on the internal and external semantic features of image-report pairs, utilizing  an image-tag recognition decoder to associate image features with their respective tags and constructing soft labels for contrastive learning to mitigate false negatives. Extensive experiments have been conducted on two downstream tasks, including five public benchmarks, demonstrating that our method outperforms other state-of-the-art methods.
\section{Methodology} \label{method}
\begin{figure}[t]
    \centering
    \includegraphics[width=\textwidth]{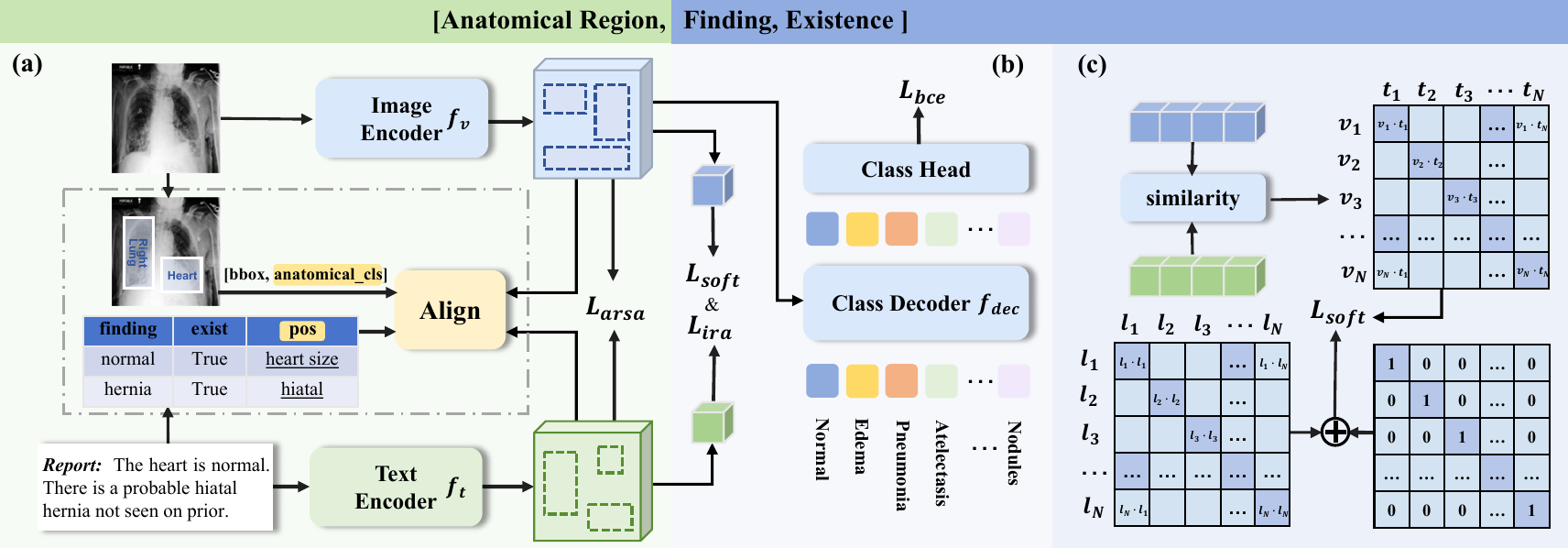}
    \caption{Overview of our ASG framework. 
    (a) Exploring local alignment between image-text pairs through anatomical region-sentence alignment. (b) Optimizing internal representation learning by applying an image-tag recognition decoder to associate image features with their respective tags. (c) Optimizing external representation learning by constructing soft labels for contrastive learning to mitigate false negatives.}
    \label{fig:framework}
\end{figure}

As illustrated in Fig.~\ref{fig:framework}, we propose a novel \textbf{A}natomical \textbf{S}tructure-\textbf{G}uided (ASG) framework for medical VLP, which consists of three parts: Image-Report Alignment (IRA), Anatomical Region-Sentence Alignment (ARSA), and Internal and External Representation Learning (IERL). In this section, we first introduce the visual and text encoding, followed by elaborating on each part in detail.

Given a batch of $N$ image-report pairs $(\textbf{I},\textbf{R})=\{(I_i, R_i)\}_{i=1}^N$, where $I_i\in\mathbb{R}^{H\times W\times 3}$ denotes the image and $R_i=\{s_i^j\}_{j=1}^{M_S}$ is the report containing $M_S$ sentences. For each image, we utilize an image encoder $f_v$ to get a sequence of $M_Z$ feature representations $\textbf{z}_i^v = f_v(I_i)\in\mathbb{R}^{M_Z\times d}$. The global visual representation is computed by averaging the dense features, \emph{i.e.}, $\textbf{v}_i$ = $\text{Avg}(\textbf{z}_i^v)\in\mathbb{R}^{d}$. For each report, we employ a text encoder $f_t$ to encode it into a sequence of sentence tokens $\textbf{z}_i^t = f_t(R_i) \in\mathbb{R}^{M_S\times d}$ and a global report feature $\textbf{t}_i\in\mathbb{R}^{d}$. 

\subsection{Image-Report Alignment (IRA)} \label{2.1}
To achieve the image-report alignment, we enforce the paired global image and report representation to be close in the feature space by employing the instance-level contrastive learning. 
Two non-linear projection layers are applied to obtain normalized lower-dimensional embeddings, \emph{i.e.}, $\hat{\textbf{v}}_i$ and $\hat{\textbf{t}}_i$.
After that, we calculate the image-to-report similarity $\textbf{p}(I_i,R_j) = \frac{\exp(\langle\hat{\textbf{v}}_i, \hat{\textbf{t}}_j\rangle/ \tau)}{\sum_{k=1}^{N}\exp(\langle\hat{\textbf{v}}_i^t, \hat{\textbf{t}}_k\rangle/ \tau)},\ 
$ and report-to-image similarity $\textbf{p}(R_i,I_j) = \frac{\exp(\langle\hat{\textbf{t}}_i, \hat{\textbf{v}}_j\rangle/ \tau)}{\sum_{k=1}^{N}\exp(\langle\hat{\textbf{t}}_i, \hat{\textbf{v}}_k^v\rangle/ \tau)},
$ where $\langle, \rangle$ represents the cosine similarity and $\tau$ is the temperature hyperparameter. 
IRA is optimized by the InfoNCE loss\cite{info} to maximize the similarity between paired instances:
\begin{equation}
\mathcal{L}_{ira}^{(v \rightarrow\, t)} = \frac{1}{N}\sum_{i=1}^{N}\mathrm{H}(\textbf{y}_i, \textbf{p}(I_i,\textbf{R})),\  
\mathcal{L}_{ira}^{(t \rightarrow\, v)} = \frac{1}{N}\sum_{i=1}^{N}\mathrm{H}(\textbf{y}_i, \textbf{p}(R_i,\textbf{I})), \\
\end{equation}
where $\mathrm{H}(,)$ denotes the cross-entropy, $\textbf{y}_i = \{(y_{ij})\}_{j=1}^N\in\mathbb{R}^{N}$ is the one-hot label with $y_{ii}$ equal to 1 and all other elements equal to 0. 
The overall objective of IRA can be denoted as $\mathcal{L}_{ira} = \frac{1}{2}(\mathcal{L}_{ira}^{(v \rightarrow\, t)} + \mathcal{L}_{ira}^{(t \rightarrow\, v)})$.

\subsection{Anatomical Region-Sentence Alignment (ARSA)}\label{2.2}
To explore fine-grained local alignment between image-text pairs, we propose an Anatomical Region-Sentence Alignment (ARSA) module by discovering clinical relevance. Specifically, we first extract the anatomical regions from the image and anatomical sentences from the report, and then we provide an automated alignment solution. Based on the aligned anatomical region-sentence pairs, we facilitate contrastive learning to realize Anatomical Region-Sentence Alignment.

\noindent \textbf{Extraction of anatomical sentences.}
For each report $R_i$, we employ RadGraph\cite{radgraph} to decompose it into a total of $M_T$ triplets $\{\mathcal{T}_j\}_{j=1}^{M_T}$, where each triplet $\mathcal{T}_j$ is denoted as <\emph{anatomical region}, \emph{finding}, \emph{existence}>, \emph{e.g.}, <\emph{lung}, \emph{pneumothorax}, \emph{exist}>. Here, the anatomical region in each triplet belongs to an anatomical set $C_{ana}$ \cite{medklip,kad}.
Notably, each triplet $\mathcal{T}_j$ corresponds to one anatomical-related sentence in the report.

\noindent \textbf{Extraction of anatomical regions.} For each image $I_i$, we use an off-the-shell Faster R-CNN \cite{fastercnn} pre-trained on Chest ImaGenome Dataset\cite{chestimg} to get a total of $M_A$ anatomical bbox $\textbf{A}_i = \{(\textbf{b}_i^j, \textbf{a}_i^j)\}^{M_A}_{j=1}$. Here, each bounding box $\textbf{b}_i^j\in\mathbb{R}^4$ represents an region with anatomical class $\textbf{a}_i^j\in C_{pre}$ in the image (\emph{e.g.}, \emph{right hilar structures}), and $C_{pre}$ is the set of pre-defined categories\cite{chestimg}.

\noindent \textbf{Automatic alignment paradigm.}
To align each anatomical bbox ($\textbf{b}_i^j, \textbf{a}_i^j$) with one triplet $\mathcal{T}$, the main challenge of building bbox-triplet alignment is two-fold. 
(1) The mismatch between the size of anatomical set $|C_{ana}|=50$ and the pre-defined categories of the detector $|C_{pre}|=29$; 
(2) The semantic overlap of $C_{ana}$ and $C_{pre}$, \emph{e.g.}, \emph{lung} defined in $C_{pre}$ is correspond to both \emph{left lung} and \emph{right lung} in $C_{ana}$.
To address the issues, we develop an automated paradigm for strict alignment based on the prior knowledge from experienced radiologists. 
In particular, given an anatomical region $pos\in C_{ans}$ and a pre-defined class $\textbf{a}\in C_{pre}$, we mainly consider three scenarios. (see Supp. Fig. 1)

\noindent\textbf{Scenario 1}: $pos$ and $\textbf{a}$ are identical words or phrases, or they share the same region despite different expressions, \emph{e.g.}, $pos$ refers to \emph{right hilar} and $\textbf{a}$ is \emph{right hilar structures}. Hence, an exact match between anatomical region $pos$ and class $\textbf{a}$ can be set.

\noindent\textbf{Scenario 2}: If \textbf{Scenario 1} is not satisfied, we pair $pos$ with an $\textbf{a}$ that can encompass it, \emph{e.g.}, $pos$ = \emph{right ventricle} and $\textbf{a}$ = \emph{cardiac silhouette}. 
Both \textbf{Scenario 1} and \textbf{Scenario 2} can be symbolized as:
\begin{equation}
\begin{aligned}
\forall j_P  \in \{1, 2, \ldots, M_T\},\ \exists j_A \in \{1, 2, \ldots, M_A\},  \ \ 
\text{s.t. } pos_i^{j_P} \rightarrow \textbf{a}_i^{j_A}.
\end{aligned}
\end{equation}

\noindent\textbf{Scenario 3}: There exists a one-to-many relationship between $pos$ and $\textbf{a}$. For example, when $pos$ is \emph{diaphragm unspec} and $\textbf{a}$ does not have bbox which can fully encompass the entire diaphragm but only includes bboxes for the \emph{left diaphragm} and \emph{right diaphragm}. 
Here, we propose two solutions: merging anatomical bboxes or splitting the sentences. 
\begin{equation}
\begin{aligned}
\forall j_P \in \{1, 2, \ldots, M_T\},\ \exists j_A, j_A'\in \{1, 2, \ldots, M_A\}, \ \ 
\text{s.t. } pos_i^{j_P} \rightarrow \{\textbf{a}_i^{j_A}, \textbf{a}_i^{j_A'}\}
\end{aligned}
\end{equation}

\begin{itemize}[label=$\bullet$]
\item Splitting the sentences - let $s_i^{j_S}$ be the matched sentence in the report for the anatomical region $pos_i^{j_P}$, we split $s_i^{j_S}$ into $s_i^{\widetilde{j}_S}$ and $s_i^{\hat{j}_S}$. Thus, the matched pairs can be formed as $(s_i^{\widetilde{j}_S}, \textbf{a}_i^{j'_A})$, $(s_i^{\hat{j}_S}, \textbf{a}_i^{j_A})$. \emph{E.g.}, \emph{diaphragm unspec} is split into \emph{left diaphragm} and \emph{right diaphragm}, ensuring a strict correspondence with two bounding boxes.\\
\item Merging anatomical bboxes - the matched pair is constructed by merging two anatomical bboxes: $(s_i^{j_S}, \textbf{a}_i^{j_A} \cup \textbf{a}_i^{j'_A})$. \emph{E.g.}, the bboxes for \emph{left diaphragm} and \emph{right diaphragm} are merged to obtain the entire diaphragm region.
\end{itemize}

After obtaining the local anatomical region-sentence pairs of each sample, we calculate the contrastive learning loss $\mathcal{L}_{arsa}$, which is optimised by InfoNCE loss implemented on the region-sentence level.

\subsection{Internal and External Representation Learning (IERL)}

In Section \ref{2.2}, we decompose the raw report into triplets and focus on anatomical structures to do the local alignment. Similarly, the finding and existence of triplets are crucial for fine-grained matching. 
We consider them as tags $\textbf{l}_i = \{l_i^j\}_{j=1}^{M_Q}$ for image-report pairs, and utilize these tags to optimize internal and external representation learning. 
If the current pair has a disease of class $j$, then $l_i^j=1$, otherwise, it is 0. Here, ${M_Q}$ is the number of disease classes. 

\noindent\textbf{Internal representation learning.} 
Internal representation learning aims to discover the relationship between image and tags within each sample.
For each image-report pair, we apply an image-tag recognition decoder $f_{dec}$ to associate image features with their respective tags. Specifically, we use the sequence of encoded visual tokens $\textbf{Z}_i = \{\textbf{z}_i^j\}^{M_Z}_{j=1}$ as both key and value, and utilize a collection of disease classes $\textbf{Q} = \{q^j\}_{j=1}^{M_Q}$ as queries. 
The classification loss is formulated as follows:
\begin{equation}
\mathcal{L}_{bce} = \frac{1}{N}\sum_{i=1}^{N}\frac{1}{M_Q}\sum_{j=1}^{M_Q}(l^j_i\log f_{dec}(\textbf{Z}_i,Q)+(1-l^j_i)\log f_{dec}(\textbf{Z}_i,Q)).
\end{equation}

\noindent\textbf{External representation learning.} External representation learning aims at improving visual-text alignment by exploring tags to connect different image-report pairs.
Previous methods\cite{kad} simply adopted hard labels by treating paired texts (reports from the same patient's study) as positives and unpaired texts (reports from other patients' studies) as negatives. Nevertheless, hard labels introduce many false negatives, as reports from different patients could have identical symptoms. 
Therefore, we explore soft label $\textbf{p}^{soft}(I_i, \textbf{R})= \textbf{p}^{soft}(R_i, \textbf{I}) = \{p_{ij}^{soft}\}_{j=1}^N $ to capture the deep semantic associations between different pairs, which are constructed based on the cosine similarity between different tags: 
\begin{equation}
p_{ij}^{soft} = \frac{\exp(\langle\textbf{l}_i, \textbf{l}_j\rangle/ \tau)}{\sum_{k=1}^{N}\exp(\langle\textbf{l}_i, \textbf{l}_k\rangle/ \tau)}.
\end{equation}

In practice, we use the weighted average of the hard labels and the soft labels as the final label to ensure the training stability and better generalization, which is formulated as $\hat{\textbf{p}}(I_i, \textbf{R})= \hat{\textbf{p}}(R_i, \textbf{I}) = (1-\alpha)\textbf{y}_i + \alpha \textbf{p}_i^{soft}.
$
Finally, the Kullback-Leibler (KL) divergence is used as the loss to minimize the distance between the final label and the similarity score $\mathcal{L}_{soft} = \frac{1}{2}(\mathcal{L}_{soft}^{(v \rightarrow\, t)}+\mathcal{L}_{soft}^{(t \rightarrow\, v)}),$ where
:
\begin{equation}
\mathcal{L}_{soft}^{(v \rightarrow\, t)} = \frac{1}{N}\sum_{i=1}^{N}\mathrm{KL}(\hat{\textbf{p}}(I_i, \textbf{R})||\textbf{p}(I_i, \textbf{R})), \ \mathcal{L}_{soft}^{(t \rightarrow\, v)} = \frac{1}{N}\sum_{i=1}^{N}\mathrm{KL}(\hat{\textbf{p}}(R_i, \textbf{I})||\textbf{p}(R_i, \textbf{I})). \\
\end{equation}
\textbf{Overall objective.}
We train our ASG framework by jointly optimizing the following losses: 
\begin{equation}
\mathcal{L} = 
\mathcal{L}_{ira} + \mathcal{L}_{arsa} + \mathcal{L}_{bce} + \mathcal{L}_{soft}.
\end{equation}
% \vspace{-3mm}

\setlength\tabcolsep{2.5pt}
\section{Experiments}
\label{Experiments}

\subsection{Experimental Setting}
\textbf{Pre-training Setting} We pre-train our framework on MIMIC-CXR\cite{mimic} and follow previous works to preprocess the dataset. The frontal view of images and the reports with more than 3 tokens are selected to generate 217k image-report pairs. We use ResNet50\cite{resnet50} and ViT-B/16\cite{vit} as the image encoders, and BioClinicalBERT\cite{biobert} is the text encoder. Our ASG is trained 50 epochs on 4 RTX 3090 GPUs with a batch size of 72 per GPU. We use AdamW\cite{adamw} as our optimizer, setting the learning rate to $4e-5$ and the weight decay to $5e-2$. And a linear warm-up and cosine annealing scheduler\cite{warmup} are applied in the process. 

\noindent \textbf{Downstream Tasks} \textbf{(1) Medical Image Classification} We conduct medical image classification on four representative datasets, NIH ChestX-ray\cite{chestxray14}, CheXpert\cite{chexpert}, RSNA\cite{rsna}, and COVIDx\cite{covid}. We use the linear probe classification setting to evaluate the transfer ability of our pre-trained image encoder. \textbf{(2) Medical Semantic Segmentation} We evaluate the performance of our framework for medical semantic segmentation on SIIM\cite{siim} and RNSA\cite{rsna} datasets. We use the pre-trained ResNet50\cite{resnet50}/ViT-B/16\cite{vit} image encoder as a frozen encoder backbone of U-Net\cite{unet}/SETR\cite{setr} and train the decoder. 

\subsection{Experimental Results}
\noindent \textbf{Results on Medical Image Classification} Notably, the final results are based on ARSA with ``merged bboxes'', with a relevant explanation being provided in Ablation Study. As shown in Table~\ref{tab:classification}, our framework achieves competitive performance across four datasets. Especially, on COVIDx with the novel disease ``COVID-19'', ASG shows significant improvements, highlighting its generalization ability. Fine-tuning with 1\% of data, our ASG outperforms MGCA by 0.6\% in AUR on NIH X-ray, demonstrating its ability to comprehend and distinguish a broader variety of diseases (see Supp. Fig. 2). Due to the prototype-level global clustering module in MGCA, it exhibits a slight advantage over our ASG on datasets with fewer categories, i.e., CheXpert and RSNA.

\begin{table}[t]
\centering
\caption{Comparison with other SOTA methods on the classification task.}
\resizebox{0.85\linewidth}{!}{
\begin{tabular}{l|ccc|ccc|ccc|ccc}
\hline
   \multirow{2}{*}{Method}
   & \multicolumn{3}{c|}{NIH X-ray (AUC)}                          & \multicolumn{3}{c|}{CheXpert (AUC)}                             & \multicolumn{3}{c|}{RSNA (AUC)}                                 & \multicolumn{3}{c}{COVIDx (ACC)}   \\
 & 1\% & 10\% & 100\%  & 1\% & 10\% & 100\%  
 & 1\% & 10\% & 100\%  & 1\% & 10\% & 100\%   \\ \hline
Random Init    & 52.1  &54.6 & 55.3  & 56.1  & 62.6  & 65.7  
               & 58.9  & 69.4 & 74.1 & 50.5  & 60.3 & 70.0       \\
ImageNet Init  & 67.0  &67.5  & 71.6 & 74.4  & 79.7 & 81.4  
               & 74.9 & 74.5  & 76.3 & 64.8  & 78.8 & 86.3       \\ \hline
\multicolumn{1}{l}{\textit{CNN-based}}     &  &  & \multicolumn{1}{c}{}  &  & & \multicolumn{1}{c}{} & & &  \multicolumn{1}{c}{} &  &      &         \\
ConVIRT \cite{convirt}    
            & 64.9 & 77.1 & 80.8  & 85.9 & 86.8 & 87.3  
            & 77.4 & 80.1 & 81.3  & 72.5 & 82.5 & 92.0  \\
GLoRIA\cite{gloria}      
            & 59.7 & 74.3 & 80.0  & 87.1 & 88.7 & 88.0  
            & 87.0 & 89.4 & \textbf{90.2}  & 66.5 & 80.5 & 88.0  \\
MedKLIP\cite{medklip}     
            & 60.9 & 74.8 & 80.1  & 82.3 & 85.4 & 87.3     
            & 83.3 & 86.6 & 88.1  & 74.5 & 83.5 & 91.3   \\
KAD\cite{kad}         
            & 78.7 & 80.7 & 82.5  & 87.2 & 88.6 & 88.7     
            & 86.7 & 88.7 & 89.9  & 73.5 & 83.0 & 90.5  \\ 
MGCA  \cite{mgca}         
            & 77.7 & 80.8 & 82.6  & 87.6  & 88.0 & 88.2
            & 87.6  & 88.6 & 89.8 & 72.0  & 83.5 & 90.5    \\ 
\rowcolor{gray!20}
Ours        & 77.0  & 81.0   & 82.9  & 87.7  & 88.2 & 88.7  
& 87.2  & \underline{88.8}   & 89.7       & 77.3  & 84.8 & \textbf{93.3} \\ \hline
\multicolumn{1}{l}{\textit{ViT-based}}      &  &  & \multicolumn{1}{c}{}  &  & & \multicolumn{1}{c}{} & & &  \multicolumn{1}{c}{} &  &      &         \\
MRM \cite{mrm}        
            & 78.0 & \underline{82.1} & 83.2  &\underline{88.5} & 88.5 & 88.7
            & 87.2  & 88.7 & 89.7 & \underline{79.0} & \underline{85.5} & \underline{92.5}    \\
MGCA \cite{mgca}       
            & \underline{78.9} & \underline{82.1} & \underline{83.5}  & \textbf{88.8}  & \textbf{89.1} & \textbf{89.7}  
            & \textbf{88.6} & \textbf{89.5} & 
        \underline{90.0}   & 74.8  & 84.8 & 92.3\\
\rowcolor{gray!20}
Ours     & \textbf{79.5} & \textbf{82.2} &\textbf{ 83.6}            & 87.9 & \underline{89.0} & \underline{89.0} 
         & \underline{88.4} & \textbf{89.5} & \textbf{90.2}   
         & \textbf{81.3}  & \textbf{87.0} & \textbf{93.3}  \\\hline

\end{tabular}
}
\label{tab:classification}
\end{table}

\begin{figure}[t]
    \begin{minipage}[]{0.5\textwidth}
        \centering
        \captionof{table}{Comparison with other SOTA methods on segmentation task.}
        \resizebox{\linewidth}{!}{
            \begin{tabular}{l|ccc|ccc}
            \hline
               \multirow{2}{*}{Method}
               & \multicolumn{3}{c|}{SIIM (Dice)}  & \multicolumn{3}{c}{RSNA (Dice)}  \\
               & 1\% & 10\% & 100\%  & 1\% & 10\% & 100\%   \\ \hline
                Random Init    & 9.00  &28.6 & 54.3  & 6.90  & 10.6  & 18.5   \\
                ImageNet Init  & 10.2  &35.5  & 63.5 & 34.8  & 39.9 & 64.0    \\ \hline
            \multicolumn{1}{l}{\textit{CNN-based}}     &  &  & \multicolumn{1}{c}{}  &  & & \multicolumn{1}{c}{}   \\
            ConVIRT\cite{convirt}   & 25.0  & 43.2 & 59.9  & 55.0 & 67.4 & 67.5  \\
            GLoRIA \cite{gloria}    & 37.4  & 57.1 & 64.2  & 60.3 & 68.7 & 68.3  \\
            MedKLIP\cite{medklip}   & 55.1  & 62.0 & 66.8  & 64.7 & 68.9 & 70.3  \\
            KAD \cite{kad}          & 58.4  & 68.2 & 69.9  & 67.9 & 68.5 & 70.3  \\ 
            MGCA  \cite{mgca}       & 49.7  & 59.3 & 64.2  & 63.0 & 68.3 & 69.8  \\ 
            \rowcolor{gray!20}
            Ours & 60.7 & 66.7 & \underline{73.6} & 68.4 & 69.9 & \underline{72.6}    \\ \hline
            \multicolumn{1}{l}{\textit{ViT-based}}     
            &  &  & \multicolumn{1}{c}{}   &  &  & \multicolumn{1}{c}{}    \\
            MRM\cite{mrm}       & \underline{68.3}  & \underline{69.5} & 72.2 &\underline{69.5}  &69.2 & 70.6   \\
            MGCA \cite{mgca}    & 60.1  & 65.4 & 69.6 &69.3  &\underline{70.0} & 72.3   \\
            \rowcolor{gray!20}
            Ours       & \textbf{71.9}   &\textbf{74.7} &\textbf{75.6} &\textbf{71.7} &\textbf{72.3} & \textbf{72.8}     \\\hline
            
            \end{tabular}
        }
        \label{tab:segmentation}
    \end{minipage}
    \begin{minipage}[]{0.45\textwidth}\label{fig:heatmap}
        \centering 
        \includegraphics[width=\textwidth]{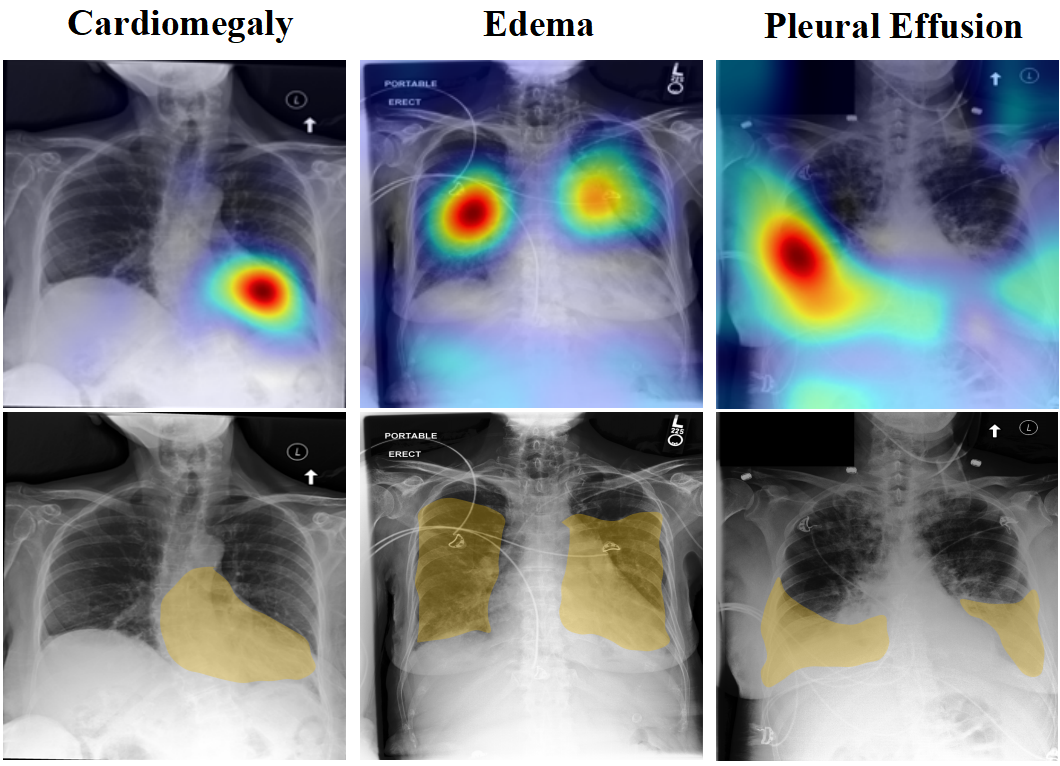} 
        \caption{Heat maps of the vision-language association learned by ASG, compared with GT annotations provided by radiologists.}
        \label{fig:heatmap}
    \end{minipage}
\end{figure}

\noindent \textbf{Results on Medical Semantic Segmentation} Table~\ref{tab:segmentation} presents the semantic segmentation performance results on the SIIM and RSNA datasets. ASG demonstrates superior performance compared to all SOTA methods across every data fractions. Notably, ASG achieves a Dice score of 71.9\% with only 1\% data fine-tuning on the smaller-scale SIIM, surpasses the runner-up method by 3.6\%, demonstrating the robust dense prediction capability of our framework.

\begin{table}[t]
\centering
\caption{Ablation study of our framework.$^{\dagger}$ and $^\#$ denotes ARSA based on merge bbox and split sentence, respectively.}
\resizebox{0.75\linewidth}{!}{
\begin{tabular}{ccc|ccc|ccc|ccc}
\hline
\multicolumn{3}{c|}{Learning Objective} & \multicolumn{3}{c|}{NIH X-ray(AUC)} & \multicolumn{3}{c|}{COVIDx(ACC)} & \multicolumn{3}{c}{RSNA(Dice)} \\
IRA &ARSA &IERL    & 1\%       & 10\%       & 100\%      & 1\%      & 10\%      & 100\%     & 1\%     & 10\%     & 100\%     \\ \hline
\checkmark   &   &   
& 78.2  & 81.7 & 82.6  & 75.3 & 85.8 & 91.0 & 65.1 & 67.7 & 68.3          \\
\checkmark   & \checkmark$^{\dagger}$   &              
& 79.1  & 81.8 & 83.1  & 77.5  & 86.0 & 92.3 & 70.6 & 71.2 & 71.9   \\
\checkmark   & \checkmark$^\#$   &              
& 78.9  & 81.5 & 83.4  & 76.3  & 86.3 & 92.0 & 69.0 & 69.4 & 69.7   \\
\checkmark   &          & \checkmark             
& 78.8  & 81.7 & 83.4  & 79.3  & 86.5 & 92.8            
& 67.4 & 68.6 & 69.7   \\
\rowcolor{gray!20}
\checkmark    & \checkmark $^{\dagger}$    & \checkmark       
&\textbf{79.5} &\textbf{82.2} &\textbf{83.6} 
&\textbf{81.3} &\textbf{87.0} &\textbf{93.3}           
&\textbf{71.7} &\textbf{72.3} &\textbf{72.8}          \\ \hline
\end{tabular}
}
\label{tab:ablation}
\end{table}

\noindent \textbf{Qualitative Analysis} As shown in Fig.~\ref{fig:heatmap}, to better understand the working mechanism of our ASG framework, we visualize the correspondence between images and disease words. ASG accurately highlights the relevant regions corresponding to a given disease, assisting the model in precise classification.

\noindent \textbf{Ablation Study} We conducted the ablation study on two tasks with three datasets, the detailed results are shown in Table~\ref{tab:ablation}.
The incorporation of the ARSA brings improvements in both classification and segmentation tasks, facilitating model to focus on local lesion representations across the entire image. ARSA based on merged bboxes outperforms that based on split sentences, likely because the former allows the model to learn the connections between different anatomical regions. The optimization of IERL is more pronounced in improving classification tasks performance, demonstrating a more reasonable approach to global representation modeling. Ultimately, the integration of these improvements yields the best overall performance.
\section{Conclusion}
\label{conclusion}
We introduce a novel Anatomical Structure-Guided framework for medical vision-language pre-training. Firstly, we parse raw reports into triplets and then utilize each element. By aligning anatomical regions and sentences, we improve the model's localization ability and interpretability. The model's capabilities are further enhanced by improvements in both internal and external representation learning. In future, we will focus on further improving sentence parsing and anatomical region extraction accuracy, for more tasks, such as report generation.
% \clearpage

%
\bibliographystyle{splncs04}
\bibliography{main}

\end{document}